\documentclass{article}

\PassOptionsToPackage{numbers, compress}{natbib}

\usepackage[final]{nips_2018}

\usepackage[utf8]{inputenc} 
\usepackage[T1]{fontenc}    
\usepackage{hyperref}       
\usepackage{url}            
\usepackage{booktabs}       
\usepackage{amsfonts}       
\usepackage{nicefrac}       
\usepackage{microtype}      
\usepackage{bm}
\usepackage{amsmath}
\title{Bayesian deep neural networks for \\low-cost neurophysiological markers of \\Alzheimer’s disease severity}

\author{
  Wolfgang Fruehwirt\textsuperscript{*, $\dagger$} \\
Medical University of Vienna  \&   \\
University of Oxford\\
   \And
    Adam D. Cobb\textsuperscript{*} \\
University of Oxford \\
  \AND
     Martin Mairhofer\\
     Medical University \\of Vienna \\
   \And  
 Leonard Weydemann\\
 Medical University \\of Vienna\\
     \And
 Heinrich Garn  \\
 Austrian Institute \\of Technology\\
     \And
 Reinhold Schmidt \\
 Medical University \\of Graz \\
   \And
 Thomas Benke \\
 Medical University of Innsbruck \\
   \And
 Peter Dal-Bianco \\
 Medical University of Vienna \\
   \And
 Gerhard Ransmayr \\
 Kepler University Hospital\\
   \And
 Markus Waser \\
 Austrian Institute \\of Technology\\
     \And
 Dieter Grossegger \\
 Dr. Grossegger\\ \& Drbal GmbH \\
   \And
 Pengfei Zhang \\
 University of Oxford \\
   \And
 Georg Dorffner   \\
 Medical University of Vienna\\
   \And
 Stephen Roberts  \\
 University of Oxford\\
}

\begin{document}

\maketitle

\begin{abstract}
As societies around the world are ageing, the number of Alzheimer's disease (AD) patients is rapidly increasing. 
To date, no low-cost, non-invasive biomarkers have been established to advance the objectivization of AD diagnosis and  
progression assessment. Here, we utilize Bayesian neural networks to develop a multivariate predictor for AD severity using a wide range of quantitative EEG (QEEG) markers.
The Bayesian treatment of neural networks both automatically controls model complexity and provides a predictive distribution over the target function, giving uncertainty bounds for our regression task. It is therefore well suited to clinical neuroscience, where data sets are typically sparse and practitioners require a precise assessment of the predictive uncertainty. 
We use data of one of the largest prospective AD EEG trials ever conducted to demonstrate the potential of Bayesian deep learning in this domain, while comparing two distinct Bayesian neural network approaches, i.e., Monte Carlo dropout and Hamiltonian Monte Carlo.

\end{abstract}

\section{Introduction}
\label{sec:Intro}

Alzheimer’s disease (AD) is the most common form of dementia and highly prevalent among elderly individuals. As societies around the world are ageing, the number of patients affected is rapidly increasing. AD is associated with an enormous disease burden regarding morbidity, mortality, and financial expenses. It impairs basic bodily functions such as walking and swallowing, has devastating impact on memory and cognition, and is ultimately fatal.
The costs of care for individuals with AD or other dementias are already enormous \cite{AlzheimerAssoc2017}, dementia being one of the costliest conditions to society \cite{Hurd2013}.

Since AD progresses over time, early accurate diagnosis and precise clinical monitoring would be essential. Yet, in daily clinical routine, AD assessment is usually done by subjective clinical interpretations once there is a strong prior that the disease is already at a progressed stage.

To date, no low-cost, non-invasive biomarkers have been established to advance the objectivization of diagnosis and disease progression assessment. For a wide use, such markers should not rely on expensive equipment (e.g., scanners for magnetic resonance imaging (MRI), or positron emission tomography (PET)), or require invasive procedures (injection of a positron emitter (PET), or lumbar puncture (analysis of cerebrospinal fluid)).
Properties like non-invasiveness, cost-effectiveness and a wide availability make electroencephalography (EEG) a potential candidate modality for such markers. Various studies have shown associations between quantitative EEG (QEEG) markers of slowing, complexity, and functional connectivity and AD (for reviews, see \cite{Dauwels2010, Jeong2004}).

Here, we utilize Bayesian neural networks \cite{neal2012bayesian} to develop a multivariate predictor for AD severity, as measured by the Mini-Mental State Examination (MMSE). We use 820 distinct QEEG markers as features, all belonging to the aforementioned categories of slowing, complexity, and functional connectivity.

Bayesian techniques allow for an adjustment of predictions through varying priors. Therefore potential variations in diagnostic settings (distinct locations) or distinct disease prevalence (varying age groups) can be accounted for. Furthermore, 
Bayesian approaches provide probabilistic predictions quantifying predictive uncertainty -- which is of utmost importance for medical practitioners. Finally, the Bayesian treatment of neural networks prevents over-fitting to sparse data sets as typically found in clinical neuroscience.

To demonstrate the potential of Bayesian deep learning techniques, we investigate the performance of two different Bayesian approaches, i.e., $(1)$ Monte Carlo dropout \cite{gal2016dropout} (MC dropout BNN) and $(2)$ Hamiltonian Monte Carlo \cite{neal2011mcmc} (HMC BNN), and compare their results to traditional (non-Bayesian) networks (NN). All models use identical QEEG features from one of the largest prospective AD EEG trials ever conducted.

\section{Materials and Methods}
\subsection{Experimental data}
\label{sec:ems}
One hundred and eighty-eight AD patients (133 with probable, 55 with possible AD diagnosis according to NINCDS-ADRDA criteria; 100 females; mean age 74.86 $\pm$8.06 standard deviation (SD); mean MMSE score 22.84 $\pm$3.70 SD; mean years of education 11.14 $\pm$3.04  SD; mean duration of illness (months) 26.88 $\pm$24.11 SD ; 115 (61.17\%) with anti-dementia medication) were considered for this investigation. They were recruited prospectively at the tertiary-referral memory clinics of the neurological departments at the Medical Universities of Graz, Innsbruck and Vienna, as well as Linz General Hospital as part of the PRODEM (Prospective Dementia Database Austria) cohort study. 
Assessment of disease severity was done using the 
MMSE \cite{Folstein1975}. Continuous EEG (alpha trace EEG recorder, 10-20 electrode placement) was analyzed for an eyes-closed (180 sec) and an eyes-open resting condition (180 sec). For details on the entire PRODEM experimental protocol and preprocessing pipeline, see \cite{Waser2016, Garn2014, Fruehwirt2016}.

\subsection{Feature generation}
 In total, 820 QEEG markers were computed considering various measures of slowing (absolute band power, relative band power, center frequency), complexity (auto-mutual information, Shannon entropy, Tsallis entropy), and functional connectivity (coherence, partial coherence, phase coherence, canonical correlation, dynamic canonical correlation, Granger causality, conditional Granger causality, cross-mutual information), as well as two resting state  conditions (eyes-closed, eyes-open), various brain regions \cite{Garn2015, Dauwels2010}, and multiple frequency bands \cite{Garn2015}.

Features were normalized and projected into a lower-dimensional space (five components) using principal components analysis, before passing the data to the models.

\subsection{Bayesian neural networks} 
In this paper we use Bayesian neural networks (BNNs) as our model for regression \cite{mackay1992practical,neal2012bayesian}. These are neural networks, where
we define the prior $p(\bm{\omega}_l)$ for each layer $l \in L$ as a product of multivariate normal distributions $\prod_{l=1}^L\mathcal{N}(\mathbf{0},\mathbf{I}/\lambda_l)$ (where $\lambda_l$ is the prior length-scale) and the Gaussian likelihood $p(\mathbf{y}\mid \bm{\omega}, \mathbf{x})$ as $\mathcal{N}(\mathbf{f}^{\bm{\omega}}( \mathbf{x}),\sigma^2)$, where $\sigma^2$ is the noise variance.
Assigning probability distributions over the weights leads to a function approximation $\mathbf{f}^{\bm{\omega}}( \mathbf{x})$ that is also a distribution. Therefore, we are also provided with important uncertainty estimates of our predictions.

In the presence of data $\{\mathbf{X},\mathbf{Y}\}$, we can perform inference to get the posterior distribution over the weights,
\begin{equation}\label{eq:Bayes}
p(\bm{\omega}\mid  \mathbf{X},\mathbf{Y}) = \frac{p(\mathbf{Y}\mid \bm{\omega}, \mathbf{X}) p(\bm{\omega})}{p(\mathbf{Y}\mid \mathbf{X})},
\end{equation}
which requires the normalizing distribution, otherwise known as the marginal likelihood, 
\begin{equation}\label{eq:ml}
p(\mathbf{Y}\mid \mathbf{X}) = \int_{\bm{\omega}} p(\mathbf{Y}\mid \bm{\omega}, \mathbf{X}) p(\bm{\omega})\mathrm{d}\bm{\omega}.
\end{equation}
The inferred posterior distribution is then used to form the predictive distribution over a new test point $\mathbf{x}^*$,
\begin{equation}\label{eq:pred}
p(\mathbf{y}^*\mid \mathbf{x}^*) = \int_{\bm{\omega}} p(\mathbf{y}^*\mid \bm{\omega}, \mathbf{x}^*) p(\bm{\omega}\mid  \mathbf{X},\mathbf{Y})\mathrm{d}\bm{\omega}.
\end{equation}

The integrations in Equations \eqref{eq:ml} and \eqref{eq:pred} are analytically intractable, which is the bottleneck in BNNs. To perform these integrations there are two common solutions. One solution is to replace the true posterior with a variational approximation that is either cheaper to sample from \cite{gal2016uncertainty} or conjugate to the likelihood \cite{mackay1992practical}. The other option is to use a Markov Chain Monte Carlo technique such as Hamiltonian Monte Carlo \cite{neal2011mcmc} to sample from the posterior.

In this work, we make a direct comparison between two BNN implementations:
\begin{enumerate}
    \item MC dropout \cite{gal2016dropout}, which is a variational approximation.
    \item Hamiltonian Monte Carlo \cite{neal2011mcmc}, which is a Markov Chain Monte Carlo technique that utilises Hamiltonian dynamics to explore the parameter space of networks.
\end{enumerate}
In addition, we also implement a standard (non-Bayesian) neural network as a useful baseline.

MC dropout is a variational approximation which implements dropout at test time of a neural network, as well as during training. 
A dropout mask is drawn from Bernoulli distributed random variables and sets a proportion of the weights in the network to zero. Dropping these weights at test time gives an approximation to the predictive distribution by sampling $T$ new dropout masks for each forward-pass through the network. This results in $T$ samples that represent the predictive distribution.

Hamiltonian Monte Carlo can be used to infer the predictive distribution in Equation \eqref{eq:pred} by introducing a Hamiltonian,
\begin{equation}
H(\bm{\omega},\mathbf{p}) = E(\bm{\omega}) + K(\mathbf{p}),
\end{equation}
that defines the total energy. This introduces an additional momentum variable $\mathbf{p}$, which is used to form the kinetic energy $K(\mathbf{p})$. However, our interest lies in the potential energy $E(\bm{\omega})$ as
\begin{equation}
E(\bm{\omega}) \propto - \log (p(\bm{\omega}\mid  \mathbf{X},\mathbf{Y})).
\end{equation}
Given this Hamiltonian, we define a distribution over the phase space $p(\bm{\omega},\mathbf{p})\propto \exp(-H(\bm{\omega},\mathbf{p}))$. Therefore in sampling from this distribution and ignoring the momentum parameters, we can retrieve the posterior distribution over $\bm{\omega}$, by using the Hamiltonian dynamics to sample from the phase space. For further details, we refer to \citet{neal2012bayesian}. 

\subsection{Metrics for comparison}

To compare results across the model we use both the mean squared error (MSE) and the standardized mean squared error (SMSE)
\cite{rasmussen:williams:2006}, where for $N$ test points, we define
\begin{equation}
    \textrm{SMSE} = \frac{1}{N} \sum_{i=1}^N\frac{(\mathbf{y}_i - \Tilde{\mathbf{f}}^{\bm{\omega}}( \mathbf{x}_i))^2}{\bm{\sigma}_i^2}.
\end{equation}
The SMSE is the MSE normalised by the predictive variance $\bm{\sigma}_i^2$ and we also define the predictive mean for a given data point $\mathbf{x}_i$ as $\Tilde{\mathbf{f}}^{\bm{\omega}}( \mathbf{x}_i)$. This is an appropriate metric for regression tasks, where we are also interested in the model predictive variance.

\subsection{Nested cross-validation and grid search}
A two-level nested cross-validation ($5\times5$) was used to determine generalization performance. We included a nested loop to perform a grid search to select the hyperparameter settings with minimal expected generalization error. For the HMC BNN, we optimized for the prior weight precision (see \cite{neal2012bayesian} for an intuition behind the weight precision) and for the MC dropout network, we optimized for dropout, which is directly related to the weight precision (see \cite{gal2016uncertainty} for details of the relationship). We also used the nested loop to optimize for the dropout rate and early stopping for the standard NN. In addition to selecting these hyperparameters in our grid search, we also searched through a range of small neural network architectures, with the smallest having a hidden layer of $100$ units to the largest having two hidden layers of $300$ units.

\section{Results and Discussion}
\label{sec:results}

For each of the five outer-fold test sets, we first selected the optimal hyperparameters for each type of network (HMC BNN, MC dropout BNN and NN) based on the performance in the five inner folds of the nested cross-validation loop. Then, we used the corresponding settings to compute the results in the outer folds, as reported in Table \ref{results}.

Overall, HMC BNNs significantly ($p$ < 0.05) outperformed MC dropout BNNs and standard NNs. Differences between model performances were assessed by statistically comparing squared errors of test set predictions.

All Bayesian models selected for the outer folds had two hidden layers of $100$ units.
The BNNs showed an average lower MSE than the best standard NN across the test sets.
This highlights the importance of combining the function approximating power of neural networks with Bayesian techniques. This is especially true for sparse data sets. In the present study, the data set consisted of $188$ patients, which allowed us to directly compute the derivatives in the Hamiltonian. As such, we saw the HMC BNN outperforming the MC dropout BNN, which is often seen in practice  \cite{gal2018idealised}. The comparably high SMSE values for the MC dropout model is due to the underestimation of the predictive uncertainty. This is a common limitation of variational inference techniques as the approximate posterior is optimised through an objective that is mode-seeking and can fail in modelling probability mass far away from the mode. In larger data sets, approximate inference (such as MC dropout) is currently the most practical way to implement BNNs.

Overall, we have shown how BNNs can be used to facilitate the development of low-cost, non-invasive markers of AD severity. To the best of our knowledge, this is the first article reporting a BNN approach for building QEEG markers of AD severity.

\tiny
\begin{table}[t]
\caption{Mean squared error (MSE) and the standardized mean squared error (SMSE) of outer loop test sets (nested cross-validation) for predicting AD severity, as measured by the Mini-Mental State Examination (MMSE) for the HMC Bayesian neural networks (HMC BNN), MC dropout Bayesian neural networks (MC dropout BNN) and standard neural networks (NN).}
  \small
  \label{results}
  \centering
  \begin{tabular}{lllllll}
    \toprule
 &  \multicolumn{2}{c}{HMC BNN} & \multicolumn{2}{c}{MC dropout BNN} & \multicolumn{2}{c}{NN} \\
    \cmidrule{2-7}
Test set      & MSE              & SMSE                   & MSE                  & SMSE                & MSE          & SMSE         \\
    \midrule
1           &  \textbf{12.42}       & \textbf{15.21}             & 17.29         &  1198.36            &23.56 & n.a.\\
2           &   14.95       & \textbf{185.41}             &\textbf{13.94}           &  1027.51         &24.94 &n.a.\\
3            &  \textbf{11.49}      &\textbf{144.14}              & 13.40         &  1030.63           &20.39 &n.a.\\
4            &   \textbf{14.36}       & \textbf{173.73}           &19.81           &  708.82           &81.21 &n.a.\\
5            &  \textbf{7.65}      &\textbf{101.62}              & 18.91          &  775.49           &18.45 &n.a.\\
                         
      \midrule
1-5            & \textbf{12.17}      & \textbf{124.02}            & 16.67          & 948.16           &33.71 &n.a.\\

    \bottomrule
  \end{tabular}
\end{table}
\normalsize

\newpage

\section*{Acknowledgements}
The PRODEM cohort study has been supported by the Austrian Research Promotion Agency FFG, project no. 827462.

\bibliography{bib}

\end{document}